\documentclass[conference]{IEEEtran}

\usepackage[letterpaper, top=19.2mm, bottom=19.2mm, left=19.2mm, right=19.2mm]{geometry}

\usepackage{cite}
\usepackage{amsmath,amssymb,amsfonts}
\usepackage{algorithmic}
\usepackage{graphicx}
\usepackage{textcomp}
\usepackage{xcolor}
\usepackage{booktabs}
\usepackage[font=footnotesize]{caption}
\usepackage{todonotes}
\usepackage{titlesec}
\usepackage{comment}
\usepackage{enumitem}
\usepackage{subfig}
\newcommand{\mytodo}[1]{\todo[inline,color=yellow!60]{\textbf{TODO:} #1}}
\def\BibTeX{{\rm B\kern-.05em{\sc i\kern-.025em b}\kern-.08em
    T\kern-.1667em\lower.7ex\hbox{E}\kern-.125emX}} 
\titlespacing{\section}{0pt}{*0.2}{*0.2}
\titlespacing{\subsection}{10pt}{*0.1}{*0.1}
\newcommand\copyrighttext{%
  \footnotesize \textcopyright 2021 IEEE. Personal use of this material is permitted.
  Permission from IEEE must be obtained for all other uses, in any current or future
  media, including reprinting/republishing this material for advertising or promotional
  purposes, creating new collective works, for resale or redistribution to servers or
  lists, or reuse of any copyrighted component of this work in other works.
  }
\newcommand\copyrightnotice{%
\begin{tikzpicture}[remember picture,overlay]
\node[anchor=south,yshift=10pt] at (current page.south) {\fbox{\parbox{\dimexpr\textwidth-\fboxsep-\fboxrule\relax}{\copyrighttext}}};
\end{tikzpicture}%
}
\begin{document}

\title{\vspace{6.3mm}\LARGE \bf Demonstrating Cloth Folding to Robots: Design and Evaluation of a 2D and a 3D User Interface }

\author{Benjamin Waymouth, Akansel Cosgun, Rhys Newbury, Tin Tran, Wesley P. Chan, Tom Drummond, \\ Elizabeth Croft \\ Monash University, Australia}
\maketitle
\copyrightnotice
\begin{abstract}
An appropriate user interface to collect human demonstration data for deformable object manipulation has been mostly overlooked in the literature. We present an interaction design for demonstrating cloth folding to robots. Users choose pick and place points on the cloth and can preview a visualization of a simulated cloth before real-robot execution. Two interfaces are proposed: A 2D display-and-mouse interface where points are placed by clicking on an image of the cloth, and a 3D Augmented Reality interface where the chosen points are placed by hand gestures. We conduct a user study with 18 participants, in which each user completed two sequential folds to achieve a cloth goal shape. Results show that while both interfaces were acceptable, the 3D interface was more suitable for understanding the task, and the 2D interface was suitable for repetition. Results also found that fold previews improve three key metrics: task efficiency, the ability to predict the final shape of the cloth, and overall user satisfaction.

\end{abstract}

\section{Introduction}

Robotic object manipulation is one of the most ubiquitous tasks throughout robotics~\cite{survey_paper}, with most works focusing on manipulating rigid objects. For robotic arms to be useful for real-world tasks, they also need to be able to manipulate deformable objects, for tasks ranging from carbon fiber layup~\cite{chanaugmented}, suturing~\cite{schulman2013generalization}, to household chores~\cite{miller2012geometric,laskey2017learning} and cloth manipulation~\cite{sanchez2018robotic}. One such task, robotic cloth folding, is of particular interest to the robotics community because of its high applicability. After early successes in robotic folding~\cite{miller2012geometric,maitin2010cloth}, the field is now dominated by deep reinforcement learning approaches~\cite{lee2020learning, wu2019learning, 9034991} where the most common approach is to perform the training in simulation before transferring the model to a real-world robot via domain randomization~\cite{matas2018sim, seita2019deep, wu2019learning, ganapathi2020learning}. Challenges for sim-to-real transfer of reinforcement learning policies are well-documented~\cite{zhao2020sim}. Sim-to-real is especially difficult for deformable objects because they can not be simulated accurately, and it is difficult to represent their state. Because of these challenges and encouraged by the successful use of expert human demonstrations in reinforcement learning~\cite{9034991, TSURUMINE201972}, research started exploring how to use demonstration data to train cloth folding policies~\cite{laskey2017learning,nair2017combining,jangir2019learning,matas2018sim}.

Existing data collection methods for human folding demonstrations include selecting grasp points in real images~\cite{laskey2017learning,8276243} or simulated images~\cite{matas2018sim}, providing a sequence of images as sub-goals~\cite{nair2017combining}, hard-coded paths pre-defined by humans~\cite{jangir2019learning}, placing physical markers on clothes~\cite{sannapaneni2017learning} and remote teleoperation~\cite{7762066}. Evidenced by the large variety of methods, it is clear that there is no standardization in the literature on how to collect human demonstration data. Furthermore, none of the existing works places a particular emphasis on the user interface for collecting demonstration data. In fact, the data collection procedures are often either omitted or only very briefly mentioned in the existing literature.


In this paper, we focus on the user interfaces for demonstrating cloth folding actions to robots. We adopt the top-down pick and place action space~\cite{lee2020learning} and give the users the freedom to choose the points. We present two interface designs: 

\begin{itemize}
    \item A 2D display-and-mouse interface where points are chosen by clicking on a live top-down stream of the cloth. Commands are provided via button clicks.
    \item A 3D Augmented Reality (AR) interface where points are chosen by hand gestures. Commands are provided via speech recognition.
\end{itemize}

Available to both interfaces, we propose previewing the simulation of a chosen fold action before executing the fold on the real robot.

\begin{figure}[t!]
    \centering
    \includegraphics[trim= 50 20 80 0, clip, width=0.49\linewidth]{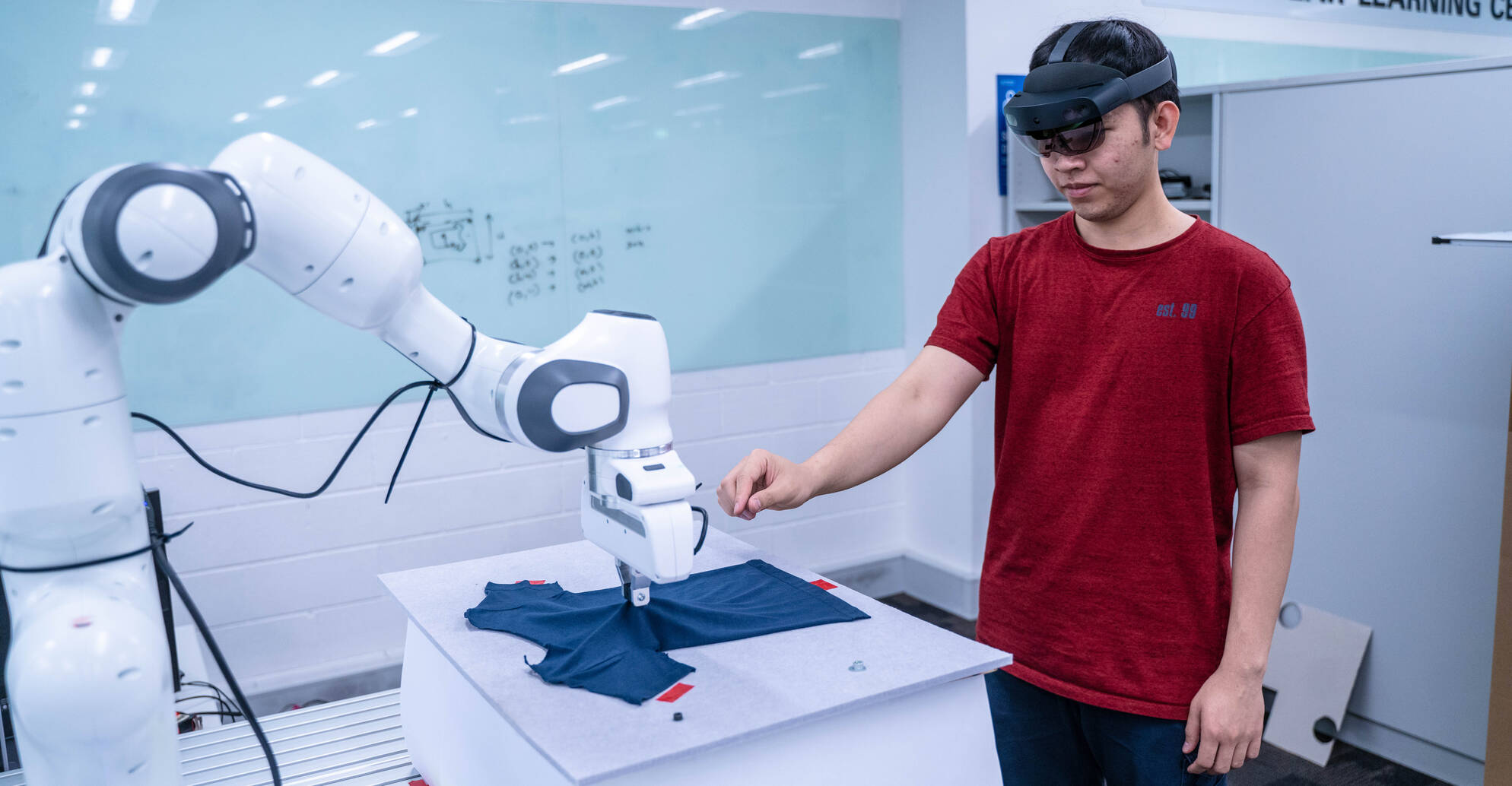}
    \includegraphics[trim= 129 0 27 0, clip, width=0.49\linewidth]{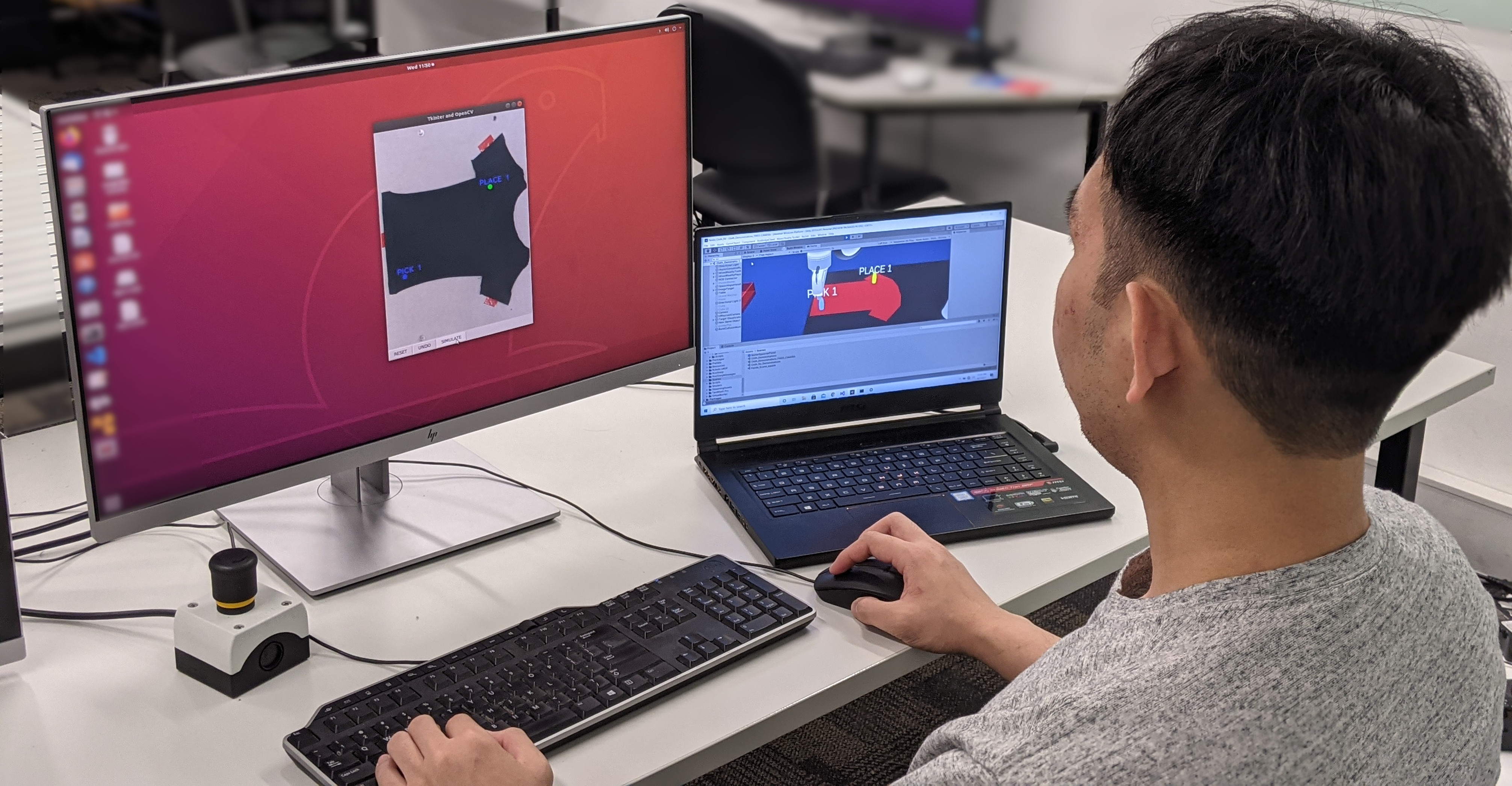}
    \includegraphics[trim= 42 65 20 65, clip, width=0.49\linewidth]{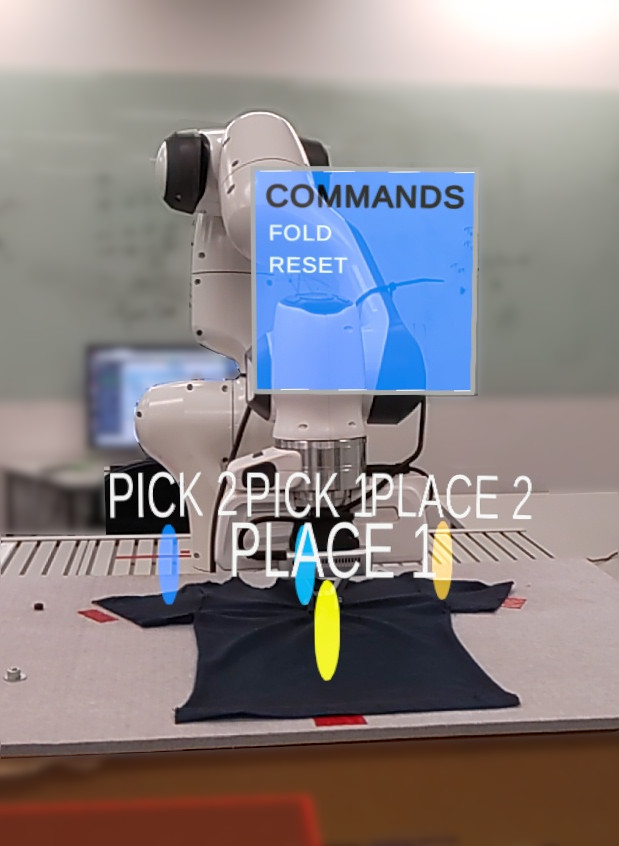}
    \includegraphics[trim= 0 0 0 50, clip, width=0.49\linewidth]{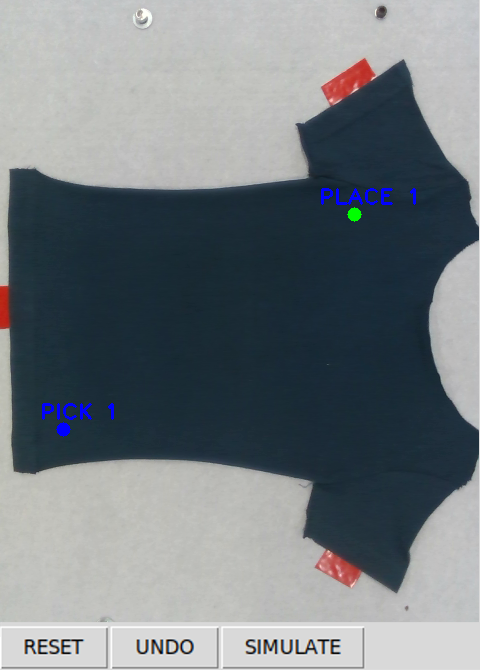}
    \caption{(Top Left) The user places virtual markers, controlling the trajectory of the robotic arm to achieve a goal configuration of the cloth (Bottom Left) The user's perspective from the HoloLens as they interacts with markers. (Top Right) Two computers are used for the 2D GUI with one screen displaying the preview of the fold (Bottom Right) The 2D GUI proposed in this paper.}
    \label{fig:intro}
\end{figure}

The contributions of this paper are two-fold:
\begin{enumerate}
    \item An interaction design for demonstrating cloth folding to robots with an option to preview fold outcomes before real robot execution, implemented on a 2D (display and mouse) and a 3D interface (Augmented Reality).
    \item User studies to evaluate the effectiveness of the proposed user interfaces and the fold preview feature for collecting fold demonstration data.
\end{enumerate}

The organization of this paper is as follows. Related literature is reviewed in Section \ref{sec:related}. The proposed system is described in Section \ref{sec:Methods}. The user study design is presented in Section \ref{sec:user_s_d}. The results are presented in Section \ref{sec:results}, before concluding in Section \ref{sec:conc}.

\section{Related Works}
\label{sec:related}

\subsection{Learning-based Cloth Manipulation}

Learning-based methods have recently dominated research in cloth manipulation, often training policies in simulation and applying sim-to-real transfer learning~\cite{matas2018sim, seita2019deep, wu2019learning, ganapathi2020learning}. All these approaches use domain randomization, where simulation parameters are sampled from a noisy estimate of the ground-truth parameters in an attempt to overcome the sim-to-real gap. However, due to the complex dynamics of cloth, there remains a large sim-to-real gap, reducing the applicability of the results. As an alternative method, human demonstrations have been shown to improve performance in learning cloth manipulation policies~\cite{matas2018sim,jangir2019learning}. 

Autonomous cloth folding is out of the scope of our work, however, our proposed interaction design is aimed at collecting high-quality human demonstration data, which typically boosts performance.

\subsection{Interfaces for Demonstrating Cloth Folding}

Various methods have been used as the user interface to collect human folding data. In \cite{laskey2017learning} users select a single grasp point by clicking on an RGB image, and an imitation learning policy is trained for bed making. Nair~\cite{nair2017combining} tackles a rope manipulation task, and human demonstrations are provided in the form of intermediate images of the rope - in this case, the robot is not involved in the data collection. Jangir~\cite{jangir2019learning} uses a hard-coded script that follows points pre-selected by a human. This interface is implemented purely in simulation and is not interactive. Matas~\cite{matas2018sim} shows that even only a handful of human demonstrations is enough to increase the folding performance, however, does not disclose how the demonstration data is collected. In~\cite{sannapaneni2017learning} users place physical round black objects on the cloths to label the grasp points. Tanaka~\cite{8276243} created a 2D Graphical User Interface (GUI) where a user could choose two grasp points and a vector representing the displacement for a two-arm humanoid robot. Yang~\cite{7762066} demonstrates folds via direct, remote teleoperation for a humanoid robot, allowing the human to control the robot using a head-mounted display. Li~\cite{li2015folding} uses simulation to allow the human to visualize the potential outcome of the folding action before being executed in the real-world. This can help overcome non-expert demonstrations impacting the model by allowing inexperienced demonstrators to optimize the demonstrations they provide.

Our 2D interface is most similar to Tanaka~\cite{8276243} and our idea to preview manipulation actions is inspired by Li~\cite{li2015folding}. We combine these ideas in a novel, user-friendly system that allows sequential folding and evaluate the system with user studies.

\subsection{Robot Programming with AR}

Due to the complex nature of programming robotic systems, many works have explored how to create intuitive robot programming interfaces. Early approaches relied on user input through auxiliary devices, such as computers or hand-held devices for defining task primitives~\cite{sbnchez2000robot, skoglund2007programming, miller2012geometric}. However, these approaches have not had much uptake. More recently, researchers have explored the benefits of using AR-based interfaces to program robots by defining low-level inputs, such as trajectory points~\cite{chanaugmented, quintero2018robot}. This process was improved by \cite{evlampev2019obstacle, ostanin2018interactive} to only require start and goal positions, allowing the robot to plan around obstacles. The work was expanded further by \cite{blankemeyer2018intuitive, wassermann2018intuitive, hernandez2020increasing} who developed high-level interfaces where users can command the robot to move an object by manipulating a virtual copy of the object.


One of the two user interfaces we explore (3D) is implemented in AR for demonstrating folds to a robot. We aim to provide the user with a more intuitive understanding of how the cloth responds to manipulation.




\section{System}
\label{sec:Methods}
\subsection{Overview}

We propose an interaction design allowing humans to demonstrate cloth folding actions to a robotic arm. We compare two interfaces: a 2D GUI and a 3D AR-based interface. Both interfaces allow the user to interact with virtual Pick and Place markers, allowing the user to control the high-level trajectory of the robotic arm. The Pick and Place markers represent top-down pick and place actions for the robot to execute. Utilizing a cloth simulator, the user has the ability to preview how the cloth would be folded when the user chooses a folding action. The user then has the option to execute the fold, undo the proposed fold or reset the virtual cloth to its initial state.





\begin{figure}[t!]
    \centering
    \includegraphics[trim= 200 100 200 0,clip,width=0.9\linewidth]{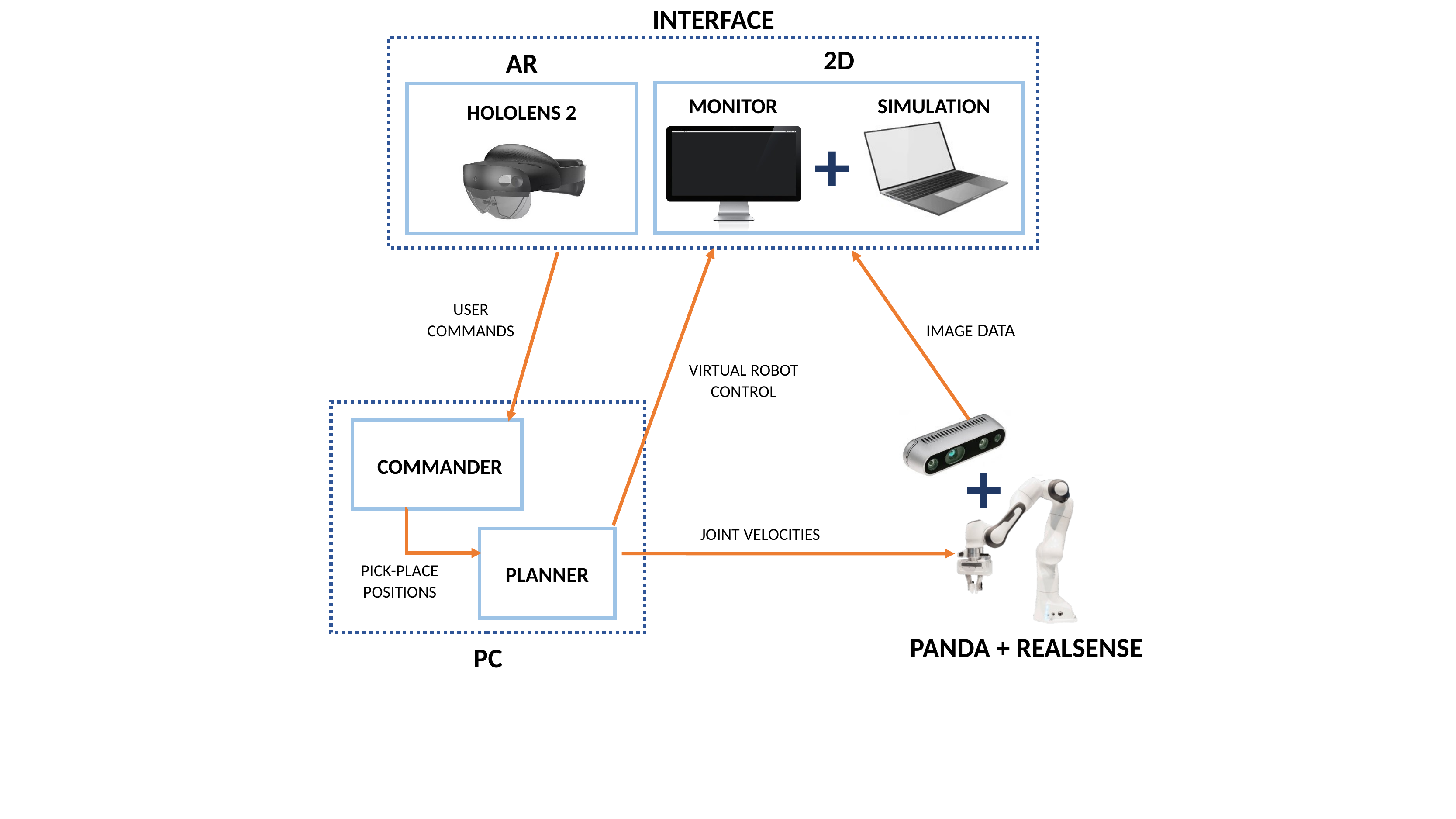}
    \caption{System Overview}
    \label{fig:system_diagram}
\end{figure}

\subsection{Hardware}

We use a table-mounted Franka Emika Panda robotic arm. The arm has 7 degrees of freedom and a parallel-jaw gripper, which can manipulate the cloth. To increase the robustness of the grasps, anti-slip tape is attached to the bottom of the gripper fingers. Our system uses a RealSense D435 RGB-D camera mounted on the end-effector. The system uses two computers, one is used for motion planning, the other is used solely for previewing the cloth folding using the 2D display. Inter-node communication is done using the Melodic Morenia version of Robot Operating System (ROS) using Ubuntu 18.04 LTS. 

We use a Microsoft HoloLens 2 for this study. The HoloLens is localized within the robot's workspace through calibration against a fixed location marker. After the initial calibration, the HoloLens can continuously localize itself against a fixed reference frame. The HoloLens also provides basic speech recognition, which gives the user quick access to basic commands. Communication between the computers and HoloLens is achieved over WiFi using a local area network.


\subsection{3D Augmented Reality Interface}

The HoloLens application is developed using Unity and MixedRealityToolKit, alongside the Obi Cloth~\cite{obi_cloth_2021} package to simulate the cloth dynamics. The interface consists of:

\begin{itemize}
    \item \textbf{Instruction Board:} A sign that displays the commands available to the user
    \item \textbf{Cloth:} A simulated cloth is shown, overlaid on the real cloth. We assume that the initial shape of the real-world cloth and its pose with respect to the robot is known.
    \item \textbf{Virtual Robotic Arm:} A simulated robot is shown, which can interact with the simulated cloth.
    \item \textbf{Interactive Markers:} The markers can be manipulated with a grabbing gesture, allowing the user to position the object in space. We constrain the markers to the cloth plane with a fixed orientation. The markers are initially placed on a virtual tray located next to the workspace. Only two interactive markers are displayed at a time, called Pick and Place, and once the user places them both, two new interactive markers are spawned for programming the next folding action.
\end{itemize}




\begin{figure}[t!]
    \centering
    \includegraphics[width=0.3525\linewidth]{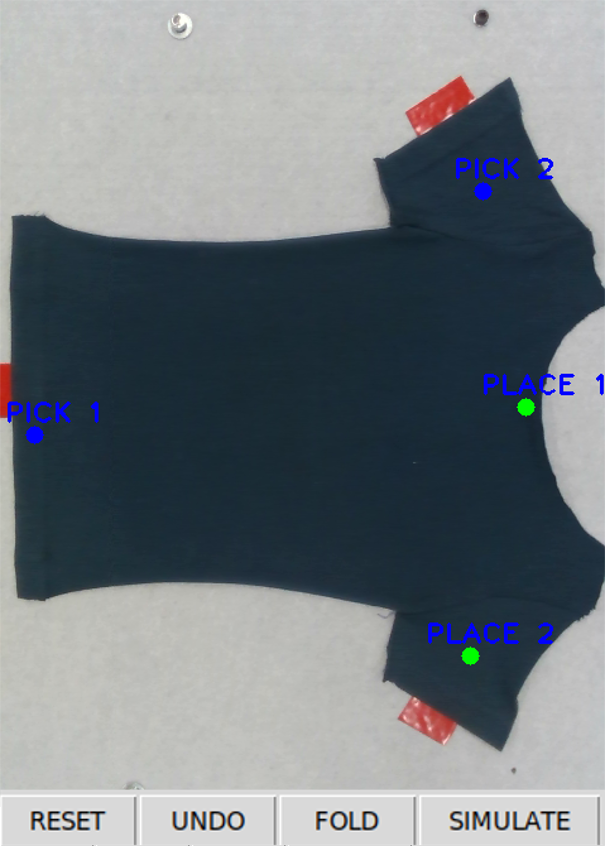}
    \includegraphics[width=0.45\linewidth]{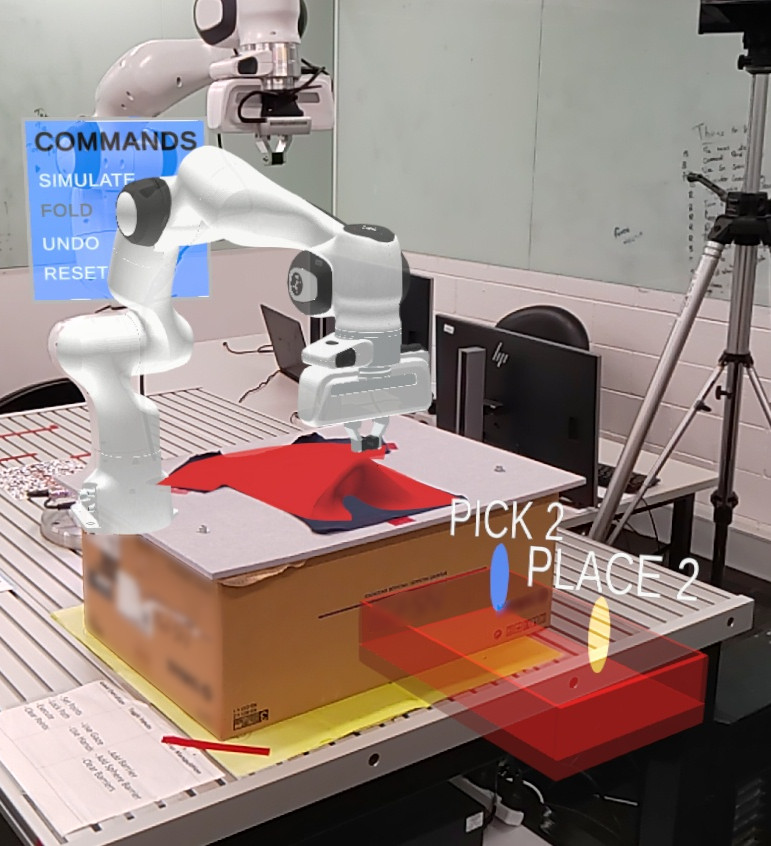}
    \caption{Left: The 2D GUI. The user has places four markers (pick and place points for two sequential folds). Commands are available with buttons at the bottom. Right: The proposed AR interface. The user has the placed first two markers and is observing the fold preview.}
    \label{fig:both_interfaces}
\end{figure}



    

\subsection{2D GUI}

The 2D GUI is shown in Fig.~\ref{fig:both_interfaces} (left image). This interface is operated via buttons (as opposed to voice commands as in the 3D AR interface). The user is presented with a live top-down view of the cloth (from the RGB-D camera attached to the robot's hand) and can select pick and place points on the image by clicking and dragging markers. The 3D coordinates of pixels belonging to the selected RGB pixels are calculated from 2D image coordinates and a depth value in a depth image. Whenever both the Pick and Place points are selected by the user, the simulation of the fold would be displayed. A second computer is used for this display.

\subsection{Interface Operation}

The experimental system was designed to keep the operation and functionalities of the 2D and AR interfaces consistent. For each interface, there are two modes: fold previews are either enabled or disabled. Each mode requires the user to define $n$ sets of Pick and Place markers, all sets of folds are performed sequentially once the user commands the robot. There is a maximum of four commands available to the user:

\begin{itemize}

\item Simulate: This command shows the user a visualization of a single fold. This consists of a virtual robot moving to the desired positions and manipulating a simulated cloth. The granularity of the cloth mesh and speed of physics had to be kept low to allow for real-time operation on the HoloLens. This was also reduced for the 2D display to ensure consistency. 

\item Fold: This command performs the two sequential folds on the real robot.

\item Undo: This command reverts the simulated cloth to the previous state to either re-watch the simulation or modify the pick and place positions.

\item Reset: This command resets all markers and the cloth (if simulation is enabled).
\end{itemize}





\subsection{Cloth Manipulation}

\begin{figure}[t!]
    \centering
    \includegraphics[trim= 0 0 0 20,clip,width=0.7\linewidth]{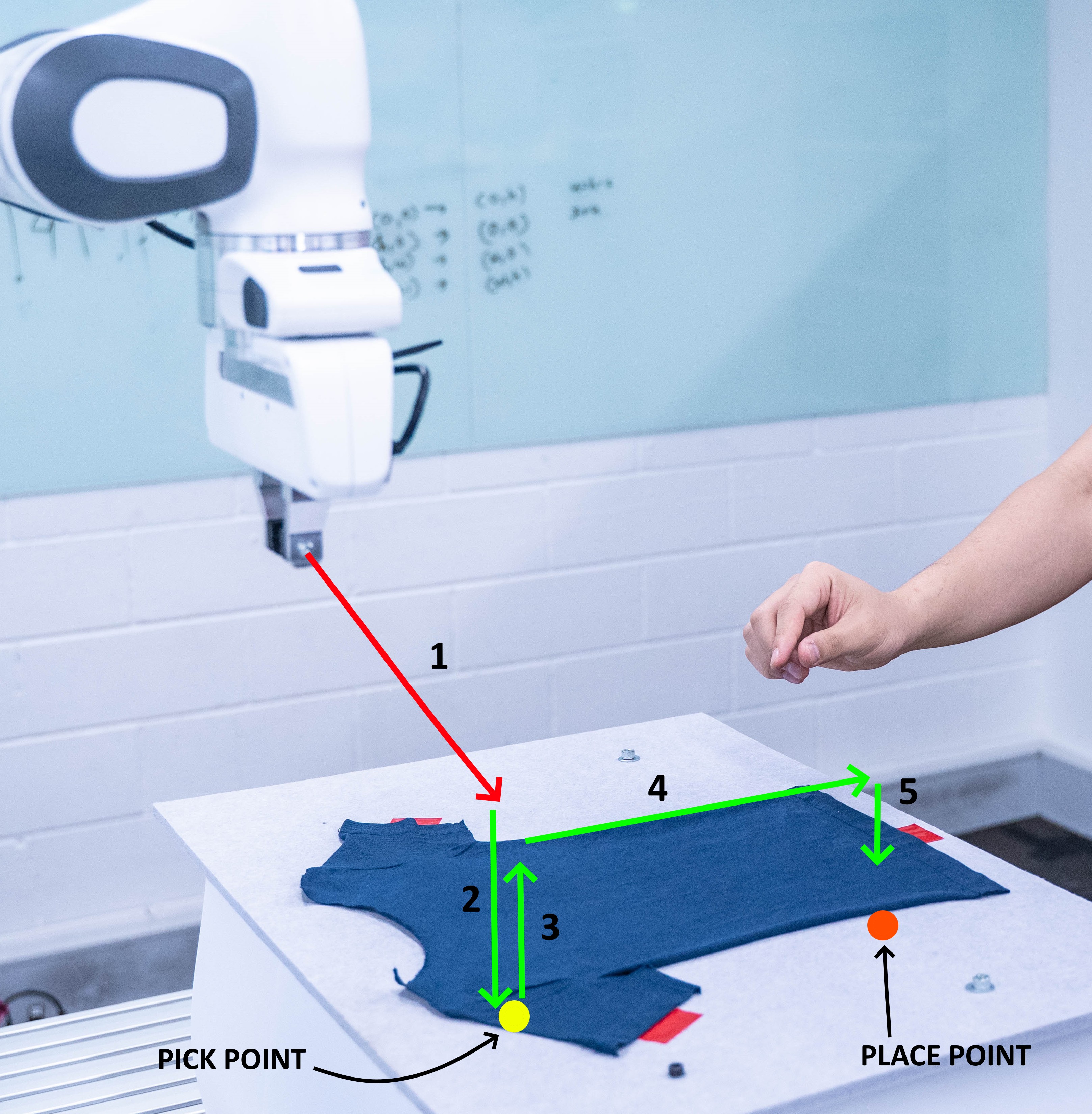}
    \caption{Each fold is executed in 5 steps: 1) Move to above the pick point. 2) Move down and pinch the cloth. 3) Move upwards to above the pick point. 4) Move above the place point at fixed height and orientation. 5) Move down and open the gripper}
    \label{fig:step_diagram}
\end{figure}

There are five manipulation steps to a folding action as shown in Figure \ref{fig:step_diagram}. Step $1$ is completed using MoveIt! to generate a trajectory. Steps $2 \rightarrow 5$ are completed using a simple inverse jacobian controller that keeps the end-effector on a linear path with fixed vertical orientation directed towards the horizontal plane. Step 1 terminates 40mm above the cloth, and the cloth is released 20mm above the plane.


\section{User Study Design}
\label{sec:user_s_d}

The study aims to explore the impact of the user interfaces and the fold preview feature when demonstrating cloth folds to a robot. The design of this study is inspired by previous work on human-robot collaboration\cite{shray}.

\subsection{Independent Variables}
We manipulate two independent variables:
\begin{enumerate}
    \item User Interface
    \begin{enumerate}
        \item \textbf{GUI}: A standard 2D computer monitor and mouse are used. The user is shown a top-down, live image of the cloth and is asked to select the pick and place points on the image by clicking and dragging.
        \item \textbf{AR Interface}: The 3D Augmented Reality interface. The user is able to select grasp and release positions on the real cloth using embodied markers and hand gestures recognized by the AR Headset.
    \end{enumerate}
    \item Fold Preview Visualization
    \begin{enumerate}
        \item \textbf{Fold Preview:} A visualization of the proposed folds is presented to the user by utilizing a cloth simulation engine before execution on the real robot.
        \item  \textbf{No Fold Preview:} Cloth folding is not visualized, hence the user has to imagine what the resultant shape of the cloth would look like, given a proposed folding action. In this mode, ``Simulate" and ``Undo" commands for both interfaces are disabled.
    \end{enumerate}
\end{enumerate}

We adopt a two by two design, therefore, each participant completed four different conditions, with a randomized order:

\begin{enumerate}
  \item \textbf{GUI, No Fold Preview}
  \item \textbf{AR, Fold Preview} 
  \item \textbf{AR, No Fold Preview}
  \item \textbf{GUI, Fold Preview}
\end{enumerate}

\subsection{Metrics}

\textbf{Survey Scores:} Five survey questions are asked to the participants after each of the 4 conditions. The questions can be seen in Table~\ref{tab:questions}. Responses were on a 7-point Likert scale.

\begin{figure}[t!]
    \centering
    \includegraphics[trim= 0 0 0 0, clip, width=0.92\linewidth]{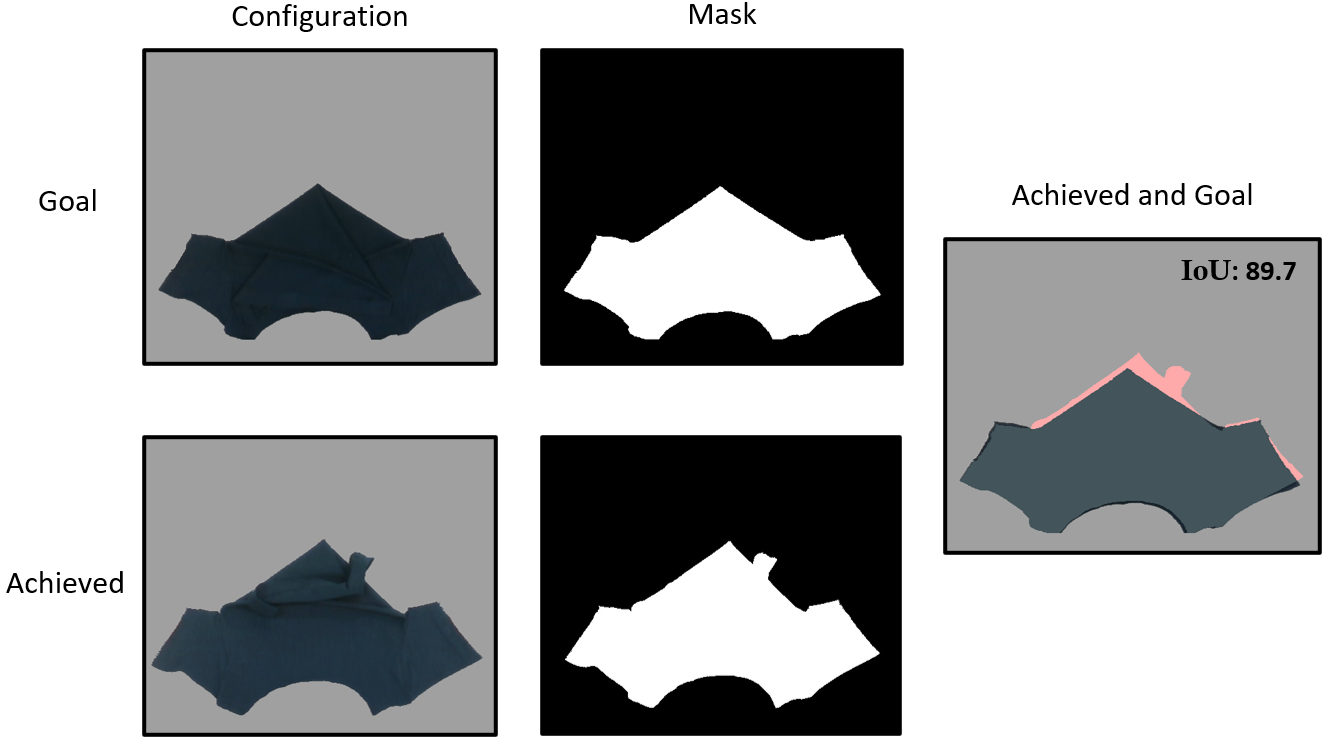}
    \caption{Fold accuracy computation using the Interface over Union (IoU) metric. On the right the achieved state is shown overlaid on the goal state, with overlapping areas shown as light blue, and error highlighted in pink.
    }
    \label{fig:IoU}
\end{figure}

\textbf{Fold Accuracy:} We use Intersection over Union (IoU) for measuring the accuracy of a fold given a cloth goal shape and an achieved cloth shape. Both shapes are captured as top-down RGB images. This metric is frequently used in computer vision community for object detection, and it has recently been proposed as a distance metric between two cloth states~\cite{lee2020learning}. See Fig. \ref{fig:IoU} for a visual description of this metric. For a given top-down view of a cloth, there are often multiple ways in which the cloth can be configured to achieve that shape, making it difficult to detect the state of the cloth, hence evaluating its similarity to a given goal configuration. For this reason, we only consider the cloth shape observed from a top-down view. All the collected goal configuration images are manually aligned in the x-y pixel space in order to compensate for the slight positional differences of the cloth when computing the IoU. The cloth masks are obtained by applying simple color segmentation in the HSV color space. A binary mask is formed from the segmented image, which is used for the IoU analysis.

\textbf{Task Completion Time:} We measure the time to complete a trial, starting when the user is given the goal configuration until the robot completes the fold and returns to the home position. 

\begin{table}[t]
    \begin{tabular}{|l|}
    \toprule
    \textbf{Q1:} How hard did you have to work to accomplish your level of \\ performance? (R) \\
    \textbf{Q2:} I was able to predict the final shape of the cloth after choosing \\ the pick and place points. \\
    \textbf{Q3:} The interface helps me complete the task efficiently. \\
    \textbf{Q4:} The interface is intuitive.  \\
    \textbf{Q5:} Overall, I am satisfied with this system. \\
    \bottomrule
    \end{tabular}
     \caption{Survey questions used in our study. (R) indicates reverse scale}
  \label{tab:questions}
\end{table}

\subsection{Participant Allocation}

We recruited 18 participants ($4$ female and $14$ male) from within Monash University.\footnote[1]{Due to the ongoing COVID-19 pandemic, no external participants could be recruited. This study has been approved by the Monash University Human Research Ethics Committee (Application ID: 27626)} Participants were aged $21-29$ (M = 23.5, SD = 2.12) and were not compensated for their time.

\subsection{Procedure}

The experiment took place at a university lab under the supervision of an experimenter. Participants first read the explanatory statement and completed a consent form. The user watched three videos explaining the study procedure and how to use each interface. 

\begin{figure}[t]
    \centering
    \includegraphics[angle=180,width=0.245\linewidth]{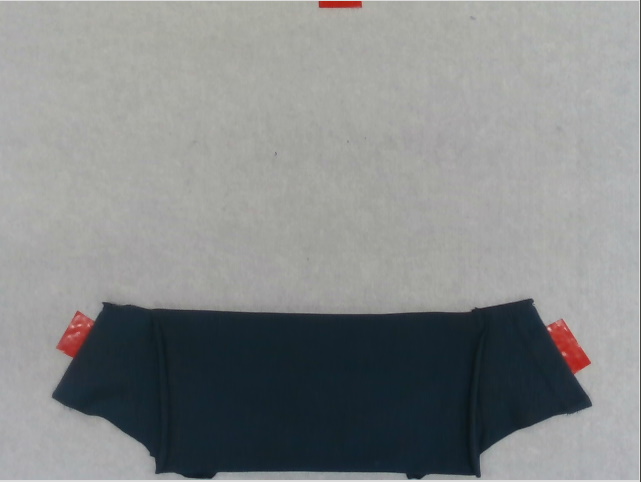}\hspace{0.1em}%
    \includegraphics[angle=180,width=0.245\linewidth]{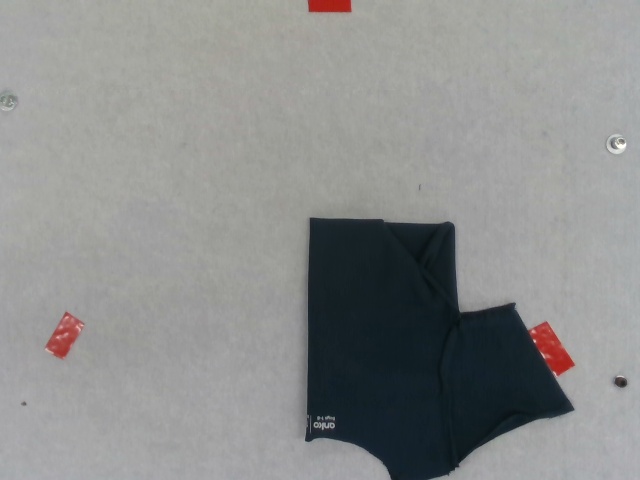}\hspace{0.1em}%
    \includegraphics[angle=180,width=0.245\linewidth]{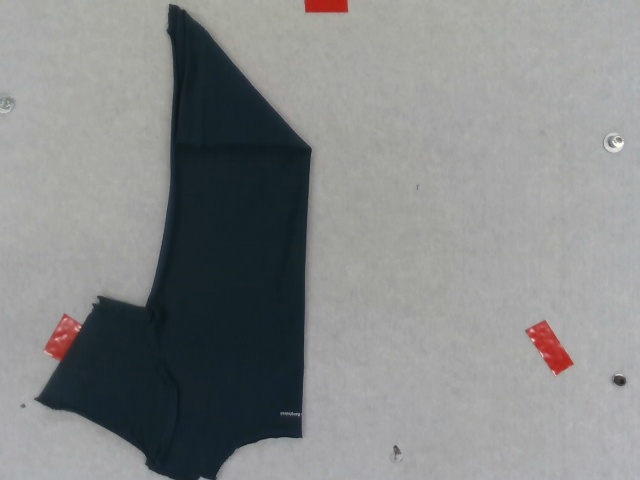}\hspace{0.1em}%
    \includegraphics[angle=180,width=0.245\linewidth]{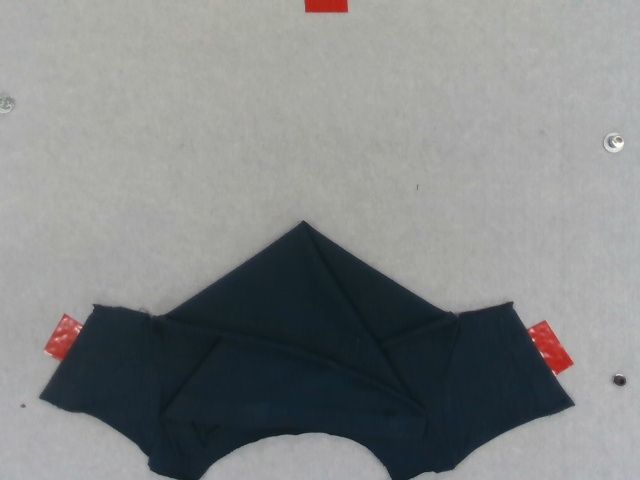}\hspace{0.1em}%
\caption{The four different possible goal configurations are provided to the user, one at a time. All goal configurations were designed to be achievable with two sequential folds.}
\label{fig:configs}
\end{figure}

The user was shown one 2D image of a folded cloth and asked to replicate the shape using the interface using exactly $n=2$ sequential folds. There were 4 possible configurations (shown in Fig.~\ref{fig:configs}) which were shown to the user in a random order, without repeating. After each trial, the participant completed a survey. This procedure is repeated over four trials. After all trials are completed, the participant was asked to provide post-study feedback, either written or oral. The average duration of an experiment was about 50 minutes. 

\section{User Study Results}
\label{sec:results}

In total we analyzed $(N = 18) * 4 = 72$ trials. 

\subsection{Choice of Interface}

We compare the two interfaces with respect to all metrics. We performed a repeated-measures two-way ANOVA. We found no significance for all questions and metrics, hence, the independent variables were collapsed. The results from these tests are shown in Table \ref{tab:interface}. We found no statistically significant difference between any of the metrics. 

\begin{table}[h!]
\setlength\tabcolsep{4pt}
\begin{tabular}{|l|c|c|c|}
\hline
   & \textbf{GUI}                     & \textbf{AR Interface}                        & p value \\ \hline
\textbf{Q1} - Hard Work  & \textbf{4.41} ($\sigma=1.9$) & 4.72 ($\sigma=1.8$)   & 0.48         \\ \hline
\textbf{Q2} - Predict Shape & 4.64 ($\sigma=1.6$) & \textbf{4.69} ($\sigma = 1.4$) & 0.88         \\ \hline
\textbf{Q3} - Task Efficiency & \textbf{5.00} ($\sigma=1.5$) & 4.69 ($\sigma = 1.9$) & 0.36        \\ \hline
\textbf{Q4} - Intuitive UI & 5.36 ($\sigma=1.1$) & \textbf{5.41} ($\sigma = 1.3$) & 0.84       \\ \hline
\textbf{Q5} - Satisfaction & \textbf{4.75} ($\sigma=1.7$) & 4.72 ($\sigma = 1.3$) & 0.92         \\ \hline
\textbf{Fold Accuracy} (\% IoU) & 73.8 ($\sigma=8.7$) & \textbf{74.9} ($\sigma = 12.5$) & 0.68 \\ \hline
\textbf{Completion Time} (s)& \textbf{240} ($\sigma=213$) & 294 ($\sigma=220$) & 0.23 \\ \hline
\end{tabular}
\caption{Results on the metrics for both interfaces. * indicates statistical significance}
\label{tab:interface}
\end{table}

The highest-scoring question for both interfaces relates to the intuitiveness of the UI, suggesting that users found both interfaces intuitive to use with no prior experience. However, the scores for both \textbf{Q1} (Hard Work) and \textbf{Q2} (Predict Shape) were lower, as folding the cloth can be a difficult task. Furthermore, the user satisfaction score was very similar, suggesting that users equally liked the two interfaces.

The mean time to complete the experiments with the 3D \textbf{AR} interface was slightly longer (294s) compared to the 2D \textbf{GUI} interface (240s). This was expected because it is more time-consuming to interact and move the pick and place markers via hand gestures to the desired position on the physical cloth compared to mouse clicking on a smaller image. The \textbf{AR} interface, however, had slightly higher goal fold accuracy (\%74.9 IoU) compared to the \textbf{GUI} interface (\%73.8 IoU).



\subsection{Effect of Having Cloth Simulation}

We use a repeated-measures two-way ANOVA test on all metrics; the results are shown in Table \ref{table:sim}. 

\begin{table}[h!]
\setlength\tabcolsep{4pt}
\begin{tabular}{|l|c|c|c|}
\hline
   & \textbf{No Fold Preview}                     & \textbf{Fold Preview}                        & p value \\ \hline
\textbf{Q1} - Hard Work & \textbf{4.22} ($\sigma=2.0$) & 4.91 ($\sigma=1.6$)   & 0.11         \\ \hline
\textbf{Q2} - Predict Shape & 4.31 ($\sigma=1.6$) & \textbf{5.02} ($\sigma = 1.4$) & *0.04         \\ \hline
\textbf{Q3} - Task Efficiency & 4.50 ($\sigma=1.5$) & \textbf{5.19} ($\sigma = 1.3$) & *0.04        \\ \hline
\textbf{Q4} - Intuitive UI & 5.25 ($\sigma=1.2$) & \textbf{5.52} ($\sigma = 1.2$) & 0.33       \\ \hline
\textbf{Q5} - Satisfaction & 4.44 ($\sigma=1.3$) & \textbf{5.02} ($\sigma = 1.1$) & *0.05         \\ \hline
\textbf{Fold Accuracy} (\% IoU) & 73.2 ($\sigma=10.7$) & \textbf{75.5} ($\sigma = 10.7$) & 0.41 \\ \hline
\textbf{Completion Time} (s) & \textbf{170} ($\sigma=134$) & 364 ($\sigma=240$) & *$<$0.01 \\ \hline
\end{tabular}
\caption{Results on the metrics for \textbf{Fold Preview} and \textbf{No Fold Preview}. * indicates statistical significance}
\label{table:sim}
\end{table}

Among the five survey questions, we find statistically significant differences in results for \textbf{Q2}, \textbf{Q3}, \textbf{Q5}. This suggests that \textbf{Fold Preview}s improve the user experience in three key metrics: the perceived task efficiency, the ability to predict the final shape of the cloth, as well as overall user satisfaction. \textbf{Fold Preview} method was also rated better in the remaining two questions regarding intuitiveness and perceived effort, however, no statistical significance was found. More user studies can provide more information about these two metrics.

The mean time to complete the experiments when the \textbf{Fold Preview} was on (364s) was significantly higher compared to when previews were off (170s). This was expected because the fold previews takes some time to compute and visualize. The longer interaction times was accompanied by better folding outcomes, evidenced by the mean fold accuracy being higher with \textbf{Fold Preview} (\%75.5 IoU) compared to \textbf{No Fold Preview} (\%73.2 IoU), showing the value of simulated fold previews.

\subsection{Qualitative Evaluation}
\label{sec:discussion}




Participant (R1) noted that the AR system was \textit{``sluggish to use, and would probably be preferred if it wasn't so slow"}. This sentiment was repeated by Participant (R9) and (R10). This behavior was due to the need to slow down the cloth simulation for it to run real-time on the HoloLens. The use of simulation tended to help people predict the final shape of the cloth (as shown by results from Q2), however, Participant (R6) indicated that they would prefer if the \textit{``simulation ran faster...as it could be frustrating waiting for it to finish"}. With more computational power, both the speed and accuracy of the simulation could be improved, allowing for faster and more accurate run times. 

Participant (R7) mentioned that they thought the click interface was superior because of the \textit{``ease of use"} and is \textit{``more inline with what I am use too"}. They also thought it was \textit{``fast"} as they could quickly click through the interface. This suggests that in practice, a good user interface does not have to be exclusively AR, or 2D.  This claim is further supported by Participant (R4) feedback on the strength and weaknesses of the two interfaces. 
\begin{itemize}

    \item \textit{``AR is good for learning"} - AR offers an intuitive, and effective method to understand the problem quickly. However, it tended to be more work physically moving the markers in space. 
    \item \textit{``2D is good for repetition"} - if we are asked to perform the same task again with small variations, 2D will offer an efficient, and \textit{``snappy"} interface to achieve a goal promptly as the user has an intuition on the problem.
\end{itemize}

The second most common negative feedback about the AR interface was the inaccuracy between the positions of the virtual pick and place markers and the actual pick and place points of the real robot (R15, R16, R17). This can be partially explained as the localization of the HoloLens is not perfect and relies on calibration relative to a fixed-location AR marker. The location of the physical cloth was assumed to be in a fixed location, however, the cloth could slightly move during the experimental setup process. Errors in pick position are also more obvious using the 3D perspective of the HoloLens, compared to the 2D interface where people have to constantly associate the approximate point on screen versus the position of the gripper.



%

%

\section{Conclusion}
\label{sec:conc}

An appropriate user interface to collect data for deformable object manipulation has been mostly overlooked in the literature, even after human demonstrations have been shown to improve the performance of learning-based cloth manipulation approaches. We present a novel interaction design for demonstrating cloth folding to robots. Users choose pick and place points on the cloth for the robot and can preview the simulated chosen fold action before the real-robot execution. This design is implemented for a 2D (display-and-mouse) and a 3D (Augmented Reality) interface.

There are pros and cons to each user interface with no clear winner. The Augmented Reality interface resulted in slightly higher fold accuracy (74.9\% IoU vs 73.8\% IoU), but at the expense of a slightly higher task completion time (294s vs 240s). Both interfaces received similar ratings in the user ratings with no statistically significant differences. The satisfaction ratings were both around 4.75 (on a scale of 7), indicating room for improvement in the user interface designs.

Fold previews were found to be highly useful for the cloth folding task in our experiments, as it was rated significantly higher than having no fold reviews in three key metrics: perceived task efficiency, ability to predict the outcome of a fold, and user satisfaction. Furthermore, with fold previews, users were able to achieve final cloth shapes closer to a given goal shape. The benefits of the fold previews comes at the expense of significantly higher task completion times. With fold previews on, the task took more than twice longer on average.


Future work can explore how to improve the user experience for this application, using the interfaces to create a dataset of cloth folding and training reinforcement learning models from human demonstration data.


\section{Acknowledgement}
This project was supported by the Australian Research Council (ARC) Discovery Project Grant DP200102858.


\bibliographystyle{IEEEtran}
\bibliography{refs}

\end{document}